\documentclass[12pt]{article}
\usepackage{epsfig}
\usepackage{chicago}
\title{Generating efficient belief models for task-oriented dialogues}

\author{Jasper Taylor}
\date{\today}

\begin{document}
\maketitle

\begin{abstract}

We have shown that belief modelling for
dialogue can be simplified if the assumption is made that the
participants are cooperating, i.e., they are not committed to any
goals requiring deception. In such domains, there is no need to
maintain individual representations of deeply nested beliefs; instead,
three specific types of belief can be used to summarize all the states
of nested belief that can exist about a domain entity.

Here, we set out to design a ``compiler'' for belief models. This
system will accept as input a description of agents' interactions with
a task domain expressed in a fully-expressive belief logic with
non-monotonic and temporal extensions. It generates an operational
belief model for use in that domain, sufficient for the requirements
of cooperative dialogue, including the negotiation of complex domain
plans. The compiled model incorporates the belief simplification
mentioned above, and also uses a simplified temporal logic of belief
based on the restricted circumstances under which beliefs can change.

We shall review the motivation for creating such a system, and
introduce a general procedure for taking a logical specification for a
domain and procesing it into an operational model. We shall then
discuss the specific changes that are made during this procedure for
limiting the level of abstraction at which the concepts of belief
nesting, default reasoning and time are expressed. Finally we shall go
through a worked example relating to the Map Task, a simple
cooperative problem-solving exercise.

\end{abstract}

\section{Introduction}

Several systems, e.g., TRAINS \shortcite{TRAINS94} have been produced to
date which model the progress of a dialogue between two agents engaged
in a common task. These systems typically represent agents as having a
set of possible reasoning methods which they can adopt in response to
a situation, e.g., they can react to what has just been said to them,
they can make a plan involving saying something themselves, or they
can attempt to recognize the plan being executed by the agent who just
spoke.

All such reasoning must take place in relation to the agents'
beliefs. A belief model capable of supporting all relevant inferences
is thus a necessary component of a simulated agent. The more complex
the reasoning being done, the more expressive the agent's belief model
must be. In particular, if agents are going to use planning and plan
recognition, they must be capable of making inferences about nested
beliefs, i.e., beliefs about other agents' beliefs.

Belief models thus correspond to epistemic logics, and are sometimes
implemented as theorem provers in standard logics, e.g.,
\citeN{wrench}. These logics allow repeated nestings of beliefs without
limit, and are computationally intractable. However, deeply-nested
beliefs appear to be useful only in equally deeply-nested reasoning,
e.g., recognizing an agent's plan to have a different plan recognized.
Reasoning at this depth of nesting is invariably associated with the attempted
practice or recognition of deception.

Furthermore, in most attempts to define logics combining epistemic and
temporal properties, e.g., \citeN{litmus,hael-for-dialogue} each
level of belief nesting includes a temporal argument, allowing the
representation of propositions such as ``Fred believed at 3pm that
Doris believed at 6pm that the cat was out''. Even with a simple
representation of time such as the Situation Calculus \cite{sit_calc},
such representations result in unnecessary complexity, as the
persistence of beliefs over time has to be deduced explicitly at each
level.

Noting that some dialogue simulators, e.g., JAM
\cite{Carletta-thesis} work well with hard limits on belief nesting
and only the simplest representation of time,
we have developed a
specification for a belief model which embodies these limits. We have
shown that as far as nesting levels are concerned, the model's
limitations cause few alterations in dialogue competence and in any
case human dialogue behaviour corresponds in some ways to that of such
a model \cite{UMUAI-paper}. However, the computational advantages of
such models can only be realized by their use in conjunction with
customized reasoning systems, in which the expressiveness of the
formulae specifying the ``ideal'' behaviour has been narrowed to cover
only the actual cases embodied by the simplified model.

\section{The compiler: an overview} \label{overview-sec}

We are currently in the process of building a ``compiler'' for belief
models. This system will accept as input a description of agents'
interactions with a task domain, expressed in a fully-expressive
belief logic with non-monotonic and temporal extensions
\cite{techreport2}. It produces a self-contained model of an agent's
beliefs and inferences in the specified domain. This model includes an
interface to events in the simulated domain based on an update
and query language which is a subset of the original fully-expressive
language. The model can thus be used in conjunction with planning and
plan recognition components to create a full simulated agent for the
domain, and multiple instances of such agents can be made to simulate
dialogue.

In overview, the process of compilation is as follows:
\begin{enumerate}
\item The fully-expressive specification is checked for constructs
  which cannot be compiled; these should not be found necessary in
  specifications of plausible domains
\item The specification is translated into a standard form resembling
  conjunctive-normal form (CNF)
\item Constructions that would create excess complexity in reasoning
  about beliefs or times are substituted for instantiations involving
  finite sets of agents, nesting depths or times
\item Clauses that contain no quantifiers are now added as premises or
  assumptions (if atomic) or justifications (if disjunctive) to an
  assumption-based truth maintenance system (DeKleer's
  \citeyear{ATMS} ATMS). The other clauses are held in a ``rule base''.
\item When an ATMS node unifies with a disjunct in a rule, a new rule
  or ATMS entity is generated by substituting the individuals in that
  node into the rule
\item Updates to the model (corresponding to ongoing events in the
  simulated domain) are translated using the same process. Queries are
  handled by examining the labels of ATMS nodes. 
\end{enumerate}

This is a fairly complex process, and will require some elucidation.
In each of the sections that follow, we shall discuss a particular
aspect of the simplifications that are made {\it en route} from the
fully-expressive specification language to the operational principles
of the compiled belief model. For each such aspect, we shall examine
the following compiler features:
\begin{itemize}
\item The restrictions applying to the specification language
\item The principles that allow reductions in complexity
\item The ways in which these principles are applied
\end{itemize}
We shall illustrate them with reference to our pilot domain, the Map
Task \shortcite{orig-maps}. This is a problem in which two agents each
have a map, and one must guide the other along a route marked only on
her own map, with reference to landmarks that may or may not be on
both maps.

\section{An adequate belief model} \label{belmod_sec}

If we work on the assumptions that agents are attempting common goals,
and that while agents may also have private goals these never require
their dialogue partners to be misled, we can eliminate the need for
explicit deeply-nested reasoning. Agents will still need to react, to
plan and to recognize plans, but in the case of planning for plan
recognition the plan to be recognized is always part of the speaker's
actual plan. So though planning for plan recognition always requires
deeply nested beliefs, under the cooperative assumption these are
always identical in content to less deeply nested beliefs.

An agent can therefore generate a full repertoire of dialogue
behaviour in a cooperative setting while distinguishing between only
three types of nesting for its beliefs, namely: beliefs about facts at
the domain level (non-nested beliefs), beliefs about the partner's
domain-level beliefs (singly nested beliefs) and beliefs about what is
believed at all deeper levels. We call the last type residual mutual
beliefs (RMB), because if two agents share all three of these types of
belief in a fact, they have mutual belief of that fact. When an agent
has evidence that the individual deeply-nested beliefs that make up a
RMB differ from each other, the RMB becomes undefined. While this is
certainly a limitation on the capabilities of the agents, it appears
that human believers in cooperative situations behave as if they have
similar limitations on their representational capabilities, i.e., they
will avoid utterances that depend for their success on such beliefs
\cite{georgia9408b,UMUAI-paper}.

Given that we are making the assumption of cooperation throughout, we
must restrict the domain specification to rules appropriate to a
cooperative domain. Specifically we disallow arbitrary implications
from one nested belief to another, e.g., ``If A believes X, then B
believes Y''. We allow the following constructs:
\begin{itemize}
\item Deduction by an agent, e.g., if Doris has a swamp on her map,
  and knows where she should be in relation to the swamp, then she
  knows where she should be on the map.
\item Relations between domain properties and beliefs, e.g., if Fred
  says he has a waterfall, then Doris believes he said this. 
\item Initial beliefs of the three types mentioned above, e.g.,
  private belief in the absence of a palm beach from one's own map,
  the presence of the route on one's partner's map, and mutual belief
  in the other relations described above. 
\end{itemize}

Even with these restrictions, there exists the potential for an
unlimited proliferation of beliefs, as a formula such as $P
\rightarrow bel(A, T, P)$ can be repeatedly unified with its own
antecedent. However, when translated into CNF, each conjunct contains
terms separated by at most one level of belief nesting. Such conjuncts
may themselves be beliefs, including mutual beliefs, in disjunctions
of belief terms, so to create the rule base the mutual belief prefixes
must be expanded into sequences that are then limited to the finite
nesting depth of the rapid model.

The rapid model itself takes the role of a single agent, so beliefs
not of that agent are irrelevant. Terms in the model can have three
belief prefixes: $obj$, meaning an object-level belief of the
simulated agent, $bel$, meaning a belief about the partner's beliefs,
and $rmb$ for residual mutual beliefs.  The $rmb$ level is only added
for beliefs that are explicitly mutual, or beliefs that recursively
imply themselves at deeper levels. So for instance, the formula
$\forall A, L, mk(utter(P) \rightarrow bel(A, utter(P)))$ (i.e., it is
mutually known that all agents know about all utterances that happen)
would be rendered $(\neg obj(utter(P)) \vee bel(utter(P))) \wedge
(\neg bel(utter(P)) \vee rmb(utter(P)))$. In the cases
where individual agents are named in the specification, e.g., it is
mutually believed that Fred does not know where the route is, beliefs
of type $rmb$ cannot be inferred, because they would have to be
about that agent's beliefs in particular and the rapid model does not
allow nesting of its belief operators.

\section{Nonmonotonicity}

There are two basic reasons for including nonmonotonic inferences in a
logic of belief for dialogue. Firstly, if the agents are having to
cooperate, then neither initially has full knowledge of the domain,
and they must rely on assumptions in order to make plans to
communicate. For instance, in the Map Task, 
agents start with the
assumption that the landmarks appearing on their own maps also apear
on their partners' maps. As the dialogue proceeds, they will
occasionally have to abandon such assumptions.

Secondly, cooperating agents use plan recognition to enhance the
effectiveness of their communication. Typically an addressee applies
plan recognition to the speaker, and then supplements her reactive
reply with an initial utterance addressing an unfulfilled goal in the
recognized plan, e.g. \citeN{trains}. Since plan recognition is
an imprecise process, and indeed most initial utterances could
potentially form part of a number of plans, the beliefs that one agent
ascribes to another in the process of plan recognition follow in part
from assumptions and must be defeasible.

Given the multiple sources of defaults in a belief model, an
appropriate logic is Hierarchical Autoepistemic Logic, or HAEL
\cite{hael-for-nonmon} which allows such defaults to be
prioritized. The specification of default reasoning in HAEL is
extremely flexible; formulae are distributed amongst evidential
spaces, which are partially ordered from strongest to weakest. A modal
operator $L_n$ allows formulae in any evidential space to refer to the
presence or absence of conclusions in stronger spaces, allowing such
concepts as ``If Wilbur is a mammal, conclude that he doesn't fly, unless
there is stronger evidence that he does'', with the stronger evidential
space containing ``If Wilbur is a bat, conclude that he flies, unless
there is stronger evidence that he does not''. 

Combining this notation with other modal operators for beliefs causes
problems; in domains such as the Map Task, the world itself is always
a certain way --- it is only the beliefs that are
uncertain. This seems to require that we use the $L_n$ operator inside
the scope of beliefs. However that would only make sense if the agent's beliefs
were themselves divided into different evidential spaces, rather than
simply occurring in them. In order to avoid this requirement we never put the
$L_n$ operator inside any modal context; fortunately, uncertain
beliefs and beliefs in uncertainties behave the same way.

To express the concept of mutual defaults, i.e., things mutually known
to be defaults, without violating this rule, we restrict the
specification language to {\em normal defaults}. These are schemata
for propositions which are always true if not contradicted. A normal
default $def_n(P)$ is equivalent to $\neg L_n(\neg P) \rightarrow P$,
where $\tau_n$ is the evidential space immediately above that
containing $def_n(P)$. For instance, in the bat example above, we
would create a normal default ``terrestrial-mammal'' ($TM$) and write
$\forall P def_1(TM(P))$, and thence $\forall X mammal(X) \wedge TM(X)
\rightarrow \neg flies(X)$. (This can be made even simpler by using
many-sorted logic and allowing $mammal$ to be a sort rather than a predicate.)

Using $def_n$, we can specify mutual defaults without putting any
references to other evidential spaces inside belief contexts. The
behaviour of a mutual default $mdef_n(P)$ is defined as if by:
$(\forall P mdef_n(P) \rightarrow def_n(P)) \wedge (\forall A, T, P
mdef_n(P) \rightarrow mdef_n(bel(A, T, P)))$. A Map Task agent's
belief that all landmarks appear on both maps unless they are known
not to, is thus a mutual default; it will hold at any level of nesting
unless specifically blocked at that level.

Constrained to these schemata, defaults can be added to the rapid
model without any further simplification. Mutual defaults are replaced
by individual default beliefs at each of the nesting levels explicitly
represented, just as mutual knowledge is replaced by individual
beliefs. The individual defaults specify the generation of assumed
nodes in the ATMS, with the assumption indicating the evidential space
in which the default holds. Where a proposition and its negation both
exist as nodes in the ATMS, the truth value of each is worked out by
looking at the labels of both to see which has justification in which
evidential space; the outcome may be:
\begin{itemize}
\item One has justification in a stronger space than the other; it is
  then true in all spaces weaker than that space, while the other is
  not true at all
\item Each is justified in a space not ordered with respect to the
  other; both are then true in different spaces
\item Both are justified in the same space; this is an inconsistency
  and should never happen.
\end{itemize}
Determining these truth values could be done by placing node states on
a bilattice, as in Ginsberg's \citeyear{MVL} logical interpretation of
the ATMS, or the pruning routine of the ATMS itself could be enhanced
to remove groups of assumptions contradicted by stronger ones, as well
as those contradicting premises.

\section{Time and reasoning about plans} \label{time_sec}

AI research in planning looks at problems that are ``nearly
decomposable''; where steps to achieve separate subgoals only
occasionally interfere with each other.  There are many temporal
formalisms available for representing such domains in logical
notations, but to model agents discussing and reasoning about such
problems, these must be combined with logics of belief. This
invariably results in the need to ascribe a time point or interval to
the belief itself as well as the believed proposition. Nested beliefs
need a time point at each level of nesting, e.g., \citeN{litmus,hael-for-dialogue}. however, the only respect in which a full
temporal logic appears to be necessary is that of the object level, where
getting temporal relations wrong might cause, e.g., trains to collide.

Beliefs themselves have temporal properties that make them easier to
model than object level facts in a complex domain. They are generated
by acts of observation or inference and persist just until some other
observation or inference overturns them. Modelling such persistence in
temporal logic requires the application of a frame axiom, and where
there are several types of event that can overturn a belief these must
be circumscriptive, i.e., be expressible as ``In the absence of any
conclusion that belief B is contradicted at time $T_1$, allow it to
persist to $T_2$''. Such axioms are also necessary to model persistence in
the domain, but because here only certain types of event can influence
a given property, circumscriptive axioms and their extra logical
complexity are typically not required \citeN{IBTL}.

The need to model the behaviour of beliefs over time is the main
motivation for the use of the ATMS. Indeed this is the motivation
behind many RMSs; we have chosen the ATMS simply because it has the
closest correspondence in its functionality to the evidential spaces
of HAEL. In its role as a continuing simulation of a believer
confronting new evidence concerning an essentially static world (i.e.,
the Map Task), the
ATMS nodes need make no reference to time at all. So our tactic in
translating formulae referring to times for use in the rapid model is
simply to strip out all temporal arguments from belief operators, and
replace all temporal relations with $true$. 

But our
agent shares her world with others, whose beliefs also change, so as
well as the case familiar in RMSs where beliefs are revised in
response to stronger evidence (as discussed by G\"ardenfors
\citeyear{kif}), beliefs about beliefs can also become out of date. For
instance, in the Map Task, one agent may at one point believe that the
other does not know where a particular landmark is, but at a later
point she may realize that from the information she has provided, the
other agent has now deduced where it is. This
would be awkward if such beliefs are added to the ATMS as premises, as there
is no way of retracting a premise, so they must be made into assumed
nodes. Each time a justification for a $\neg bel(\dots)$ or
$\neg rmb(\dots)$ node 
is added, a special assumption is included in its
antecedents. This assumption includes the ordinality of the time at
which it is added, so when labels justifying contradictory beliefs are
compared, that depending on the most recent evidence can be preferred.

When we extend the system to cope with domains in which physical
change occurs, such as the TRAINS world, we will need a more complex
representation of temporal interactions. \citeN{TWEAK} points
the way for doing temporal reasoning in a network structure; the truth
of a proposition at a given time in his planner TWEAK is defined by a
{\em modal truth criterion (MTC)}. \citeN[Ch.\ 4]{Taylor-thesis}
describes how a simplified version of Chapman's MTC can be used in
conjunction with the ATMS to model an agent's estimation of the value
of a proposition at a given time as knowledge of relevant events at
earlier times is added to the agent's world knowledge. Here, the
system is also used for keeping track of other agents' beliefs, a task for
which it is overcomplex. Such a system
in conjunction with the currently proposed method of tagging
out-of-date beliefs should result in a highly efficient system for
reasoning about plans.

\section{A simple illustration}

The compiler itself is currently under construction. So far, we have
developed a characterization of the Map Task supporting the same
dialogues as Carletta's JAM \citeyear{Carletta-thesis} as a test
domain. This characterization has been formalized in our
fully-expressive language, and we have completed a package in Prolog
for carrying out the first stage of its conversion to to the rapid
model, i.e., conversion into CNF. The resulting characterization can
be analysed with a theorem prover, also written in Prolog, to
illustrate that the desired inferences all follow, although quite
understandably, the computational complexity of actually following a
whole dialogue by this means is prohibitive.

In parallel with this exercise we hace constructed a rapid model for
the same task, based on an ATMS and a rule-based inference system as
has been described for the models generated by the compiler. This was
originally developed simply with the object of constructing a working
system to support Map Task dialogue simulation \cite{techreport1}. For
the purposes of the work described here,
it was compared with the logical characterization of the domain, and
then both were refined to the point where an orderly relationship
existed between the formulae of one and the node names and inference
rules of the other. This relationship conformed to the overview of the
compilation process given in section~\ref{overview-sec}, and is the
basis for the automated translation from one to the other which is the
subject of work at the time of writing. When this is complete, the
next step will be to specify the update and query language; this will
be a simplified version of the fully-expressive language used to
create the model, eliminating modal operators except when applying to
whole formulae.

To illustrate the compilation process as it is curently envisaged, we
will take just a few axioms from the Map Task; those required for one
agent to believe that after the other has given him a description of
part of the route, he believes the other agent to believe as a default
that he is vividly acquainted with that route section. We use
many-sorted notation, for the practical advantages noted by \cite{many-sorts} and because it allows time and agent variables to be
marked for the special preprocessing they require. Informally, the
language describes the domain as follows: terms can be of sort $agent,$
$time,$ $landmark,$ $route\_section,$ $map\_item,$ $mood$ (of an
  utterance), $relation$ or $prop$ (standing for propositions in
quantified modal formulae.) The sorts
$landmark$ and $route\_section$ are subsorts of $map\_item$. We then have the following predicates to
comprise what agents can know about the domain:
\vspace{0.25in}
\begin{center}
\begin{tabular}{|l|l|}
\hline
vivid(L)        &       Equivalent to the conjunction of predicates \\
                &       defining where L is (always true at domain level) \\
mapped(L)       &       Equivalent to the conjunction of predicates \\
                &       asserted by the perception of L on the map \\
                &       (only makes sense within belief scope) \\
desc(L, R, M)   &       Landmark L is in relation R to landmark M \\
say(A, T, M, P) &       Agent A at T voiced proposition P with mood M \\
\hline
\end{tabular}
\end{center}
\vspace{0.25in}

\subsection{The relevant axioms in the fully-expressive language}

The axioms are as follows:

Map information makes landmarks vivid
\begin{equation} \label{mpcv_eqn}
\forall X {:} map\_item \; mk(mapped(X) => vivid(X))
\end{equation}

Utterances always get heard:
\begin{displaymath}
\forall X {:} agent, Y {:} agent, Z {:} time, W {:} time, V {:} mood,
U {:} prop \; 
\end{displaymath}
\begin{equation} \label{percep_eqn}
mk(say(X, Z, V, U) \wedge before(Z, W) => 
                bel(Y, W, say(X, Z, V, U)))
\end{equation}

All assertions are true (Our model assumes the planner treats truth as a
precondition for utterances; this could be done differently)
\begin{equation} \label{truth_eqn}
\forall X {:} agent, Y {:} time, W {:} prop \; mk(say(X, Y, assert, W) => W)
\end{equation}

Descriptions confer vividness
\begin{displaymath}
\forall X {:} map\_item, Y {:} relation, Z {:} map\_item \; \end{displaymath}
\begin{equation} \label{dcv_eqn}
mk(vivid(Z) \wedge desc(X, Y, Z) 
        => vivid(X))
\end{equation}

All landmarks (but not all map items) are expected to be on other
agents' maps
\begin{displaymath}
\forall X {:} agent, Y {:} agent, Z {:} time, W {:} landmark
        \; \end{displaymath}
\begin{equation} \label{share_eqn}
mdef_{20}(bel(X, Z, bel(Y, Z, mapped(W))))
\end{equation}

Route sections are known to be on the instruction giver's
(Doris') map
\begin{displaymath}
\forall W {:} agent, X {:} route\_section, Y {:} time, Z {:} time \; \end{displaymath}
\begin{equation} \label{IG_know_eqn}
mk(bel(W, Y, bel(doris, Z, mapped(X))))
\end{equation}

\subsection{Translation into rules for driving the ATMS}

These are first translated into CNF, and then fitted to the
requirements of the rapid model. This involves the expansion of mutual
beliefs and mutual defaults into finite series of individual beliefs
and defaults about beliefs as described in
section~\ref{belmod_sec}. Time parameters are simply stripped out, as
there is currently no use for them in the domain. The results are as
follows:

\ref{mpcv_eqn} becomes:
\begin{equation} \label{mpcv_eqn_r1}
\forall X {:} map\_item \; obj(\neg mapped(X) \vee vivid(X))
\end{equation}
\begin{equation} \label{mpcv_eqn_r2}
\forall X {:} map\_item \; bel(\neg mapped(X) \vee vivid(X))
\end{equation}
\begin{equation} \label{mpcv_eqn_r3}
\forall X {:} map\_item \; rmb(\neg mapped(X) \vee vivid(X))
\end{equation}

\ref{percep_eqn} becomes:
\begin{equation} \label{percep_eqn_r1}
\forall V {:} mood, U {:} prop
        \; \neg obj(say(doris, V, U)) \vee bel(say(doris, V, U)))
\end{equation}
\begin{equation} \label{percep_eqn_r2}
\forall V {:} mood, U {:} prop
        \; \neg bel(say(doris, V, U)) \vee rmb(say(doris, V, U)))
\end{equation}
A similar pair relating to Fred are also created. Note that the
original implication from a shallow nested belief to an identical
deeper one allows the $rmb(\dots)$ to be included.

\ref{truth_eqn} becomes:
\begin{equation} \label{truth_eqn_r1}
\forall W {:} prop \; obj(\neg say(doris assert, W) \vee W)
\end{equation}
\begin{equation} \label{truth_eqn_r2}
\forall W {:} prop \; bel(\neg say(doris assert, W) \vee W)
\end{equation}
\begin{equation} \label{truth_eqn_r3}
\forall W {:} prop \; rmb(\neg say(doris assert, W) \vee W)
\end{equation}

\ref{dcv_eqn} becomes:
\begin{displaymath}
\forall X {:} map\_item, Y {:} relation, Z {:} map\_item \; \end{displaymath}
\begin{equation} \label{dcv_eqn_r1}
obj(\neg vivid(Z) \vee \neg
desc(X, Y, Z) \vee vivid(X)))
\end{equation}
\begin{displaymath}
\forall X {:} map\_item, Y {:} relation, Z {:} map\_item \; \end{displaymath}
\begin{equation} \label{dcv_eqn_r2}
bel(\neg vivid(Z) \vee \neg
desc(X, Y, Z) \vee vivid(X)))
\end{equation}
\begin{displaymath}
\forall X {:} map\_item, Y {:} relation, Z {:} map\_item \; \end{displaymath}
\begin{equation} \label{dcv_eqn_r3}
rmb(\neg vivid(Z) \vee \neg
desc(X, Y, Z) \vee vivid(X)))
\end{equation}

\ref{share_eqn} becomes:
\begin{equation} \label{share_eqn_r1}
\forall W {:} landmark
        \; def_{20}(bel(mapped(W)))
\end{equation}
\begin{equation} \label{share_eqn_r2}
\forall W {:} landmark
        \; def_{20}(rmb(mapped(W)))
\end{equation}

\ref{IG_know_eqn} becomes:
\begin{equation} \label{IG_know_eqn_r}
\forall X {:} route\_section \; bel(mapped(X))
\end{equation}
This is the only rule produced for this formula; although the original
refers to all levels below the first, putting $rmb(mapped(X))$ here
would refer to Doris' beliefs about Fred's map.

\subsection{Operation of the ATMS model}

All the above contain quantifiers, so none of them cause any nodes to
be added initially to the ATMS. Formulae corresponding to map
information are also included; these will be added using the
update and query language. Suppose Fred has a swamp on his map, which he
refers to internally as $swamp\_1$. His database is initially updated
with the proposition $obj(mapped(swamp\_1))$ which, containing no
quantifiers, is added as a premise in the ATMS. Once in the ATMS, a
test is made to see if it resolves with any disjunct from the rule
base; it does with \ref{mpcv_eqn_r1}, and the substituted version of
this is added to the ATMS as a justification from $obj(mapped(swamp\_1))$
to $obj(vivid(swamp\_1))$, the latter being supported by this
justification.

Now suppose Fred's partner Doris opens the dialogue by saying ``The
first part of the route goes left of the palm beach''. As we are looking
at a database that is modelling Fred, his hearing this utterance is
modelled by his awareness of it being established by its assertion in
the update and query language: $obj(say(doris, assert, desc(section\_1, left\_of,
palm\_beach\_1)))$. This too is unquantified, so produces a premise node
in the ATMS. It unifies with \ref{percep_eqn_r1} to produce
$bel(say(doris, assert, desc(section\_1, left\_of, palm\_beach\_1))$,
which in turn unifies with \ref{percep_eqn_r2} justifying \\
$rmb(say(doris, assert, desc(section\_1, left\_of,
palm\_beach\_1))$. These three unify with terms in \ref{truth_eqn_r1} --
\ref{truth_eqn_r3} to justify beliefs at the three levels in the actual
description, which in turn unify with \ref{dcv_eqn_r1} --
\ref{dcv_eqn_r3} to produce justifications for vividness of the first
section based on vividness of the palm beach.

However the nodes $obj(vivid(palm\_beach\_1))$,
$bel(vivid(palm\_beach\_1))$ and $rmb(vivid(palm\_beach\_1))$ as yet have
no justification. They do however unify with \ref{mpcv_eqn_r1} --
\ref{mpcv_eqn_r3}, giving them justifications from
$obj(mapped(palm\_beach\_1))$,  $bel(mapped(palm\_beach\_1))$ and
$rmb(mapped(palm\_beach\_1))$, and the latter two of these match up to
the default clauses in \ref{share_eqn_r1} and \ref{share_eqn_r2},
which result in assumptions being created for
them. $bel(mapped(palm\_beach\_1))$ and $rmb(mapped(palm\_beach\_1))$
each get a $def_{20}$ assumption, which is created with a serial
number. 
These assumptions
are propagated through the links that have been added, and appear in the
labels of the nodes for $bel(vivid(section\_1))$ and
$rmb(vivid(section\_1))$. The first of these is made a premise by
\ref{mpcv_eqn_r2} and \ref{IG_know_eqn_r}, and the assumption label
for this becomes redundant and is removed by the ATMS, but the second
remains as an assumption.

\begin{figure}
\epsfig{file=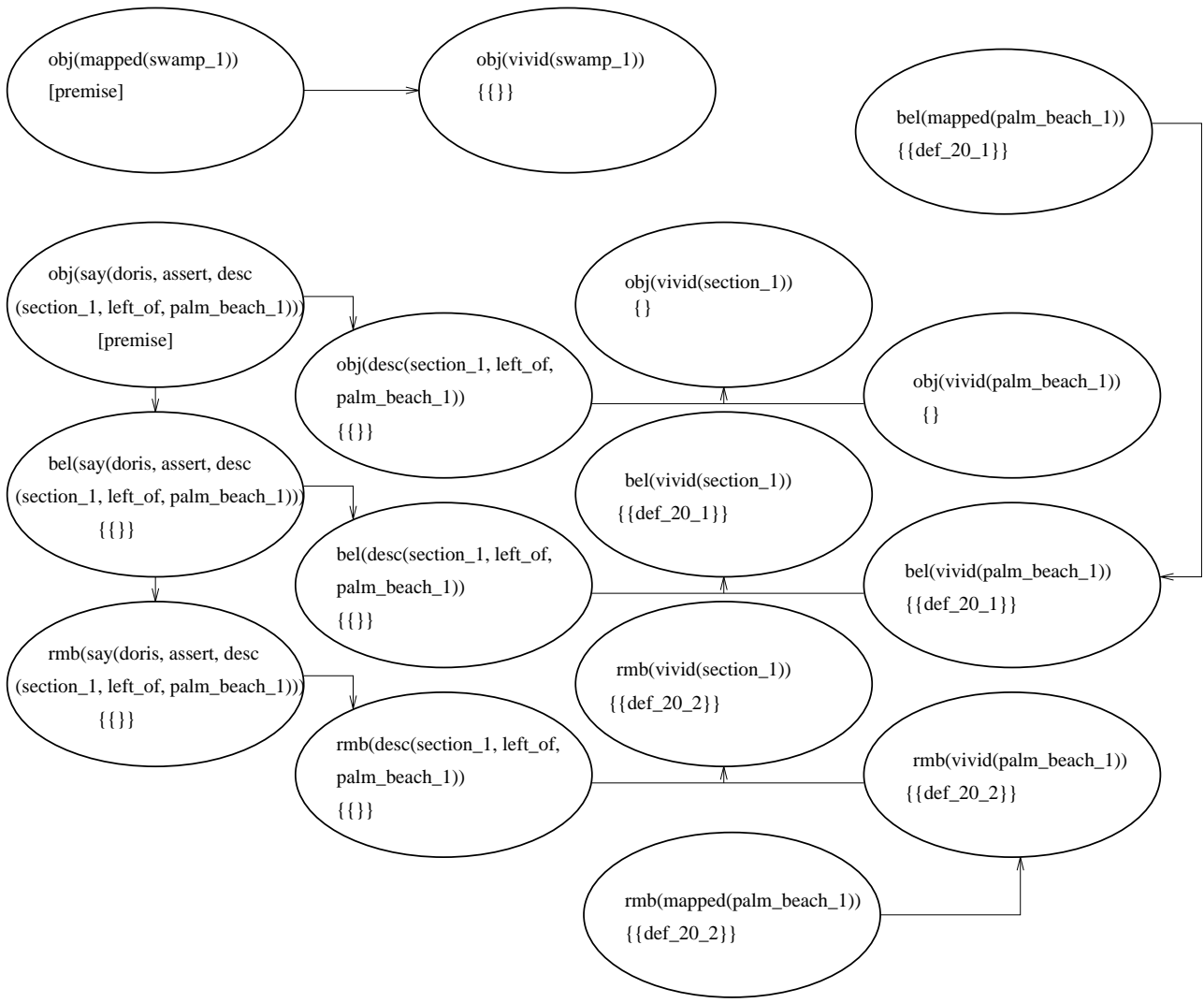,width=400pt}
\caption{Illustration of the ATMS model of Fred's belief state}
\label{ATMS_state_fig}
\end{figure}

This is the situation illustrated in figure~\ref{ATMS_state_fig}. What
happens next is that Fred makes an utterance indicating he has not
understood the description; this will give him a justification for
$\neg rmb(vivid(section\_1))$ depending only on a current time-stamp
assumption. This will form a contradiction with the positive node and
result in the assumption justifying it ($def_{20}\_2$) being removed
from the system. Subsequently, queries to the database regarding the
status of $rmb(vivid(section\_1))$ will be answered in the negative.
In her next utterance, Doris may offer Fred a description of the palm
beach in terms of the swamp, which he has (and which she can assume
him to have). This will create a new justification for
$rmb(vivid(section\_1))$, resulting in the first section finally
becoming mutually vivid for them.

\section{Conclusion}

The compiler should illustrate the adequacy of belief models of the
type that we propose, by allowing the creation of a number of such
models and their comparison with the results that would be obtained
directly from the original specification of the domain, as well as the
performance of systems using these models in conjunction with planning
and plan recognition systems to model conversational agents.

Furthermore it is expected that the models produced by the compiler in
this way will have sufficiently rapid performance to allow their use
in real-time dialogue systems. This is the case for a prototype model
developed for a Map Task dialogue system, whereas
use of an automated theorem prover working in the unrestricted logical
language in this domain appears to be impossible due to computational
complexity.

\end{document}